\documentclass[runningheads]{llncs}
\usepackage{wrapfig}
\usepackage[table,xcdraw]
{xcolor}
\usepackage{graphicx}
\usepackage{dsfont}
\usepackage{multirow}
\usepackage[pagebackref,breaklinks,colorlinks,bookmarks=false]{hyperref}

\usepackage{enumitem}
\usepackage{tikz}
\usepackage{comment}
\usepackage{amsmath,amssymb} % define this before the line numbering.
\usepackage{booktabs}
\usepackage{graphicx}
\usepackage{bm}
\DeclareMathOperator{\Tr}{Tr}
\definecolor{Gray}{gray}{0.95}
\definecolor{highlight}{RGB}{230, 255, 225}
\usepackage[font=small,labelfont=bf]{caption}

\usepackage{cleveref}
\Crefname{equation}{Eq.}{Eqs.}
\Crefname{figure}{Fig.}{Figs.}
\Crefname{tabular}{Tab.}{Tabs.}
\Crefname{section}{Sec.}{Secs.}
\Crefname{table}{Tab.}{Tabs.}
\Crefname{algorithm}{Algo.}{Algo.}

\usepackage{xspace}

\usepackage{algorithm}
\usepackage[noend]{algpseudocode}

\algdef{SE}[DOWHILE]{Do}{doWhile}{\algorithmicdo}[1]{\algorithmicwhile\ #1}%
\newcommand{\cls}[1]{{\small\texttt{#1}}}

\newcommand{\etal}{\emph{et al}.}

\begin{document}
% \linenumbers
\pagestyle{headings}
\mainmatter
\def\ECCVSubNumber{3984}  % Insert your submission number here
\newcommand{\gcdwfshort}{NCDwF\xspace}

\newcommand{\gcdwf}{Novel Class Discovery without Forgetting\xspace}

\newcommand{\LII}{Known Class Identifier\xspace}
\newcommand{\LIIshort}{KCI\xspace}

\title{Novel Class Discovery without Forgetting}

\begin{comment}
\titlerunning{ECCV-22 submission ID \ECCVSubNumber} 
\authorrunning{ECCV-22 submission ID \ECCVSubNumber} 
\author{Anonymous ECCV submission}
\institute{Paper ID \ECCVSubNumber}
\end{comment}
%******************

% \begin{comment}
\titlerunning{Novel Class Discovery without Forgetting}
\author{K J Joseph\inst{1, 2} \and
Sujoy Paul\inst{1} \and
Gaurav Aggarwal\inst{1} \and
Soma Biswas\inst{3} \and
Piyush Rai\inst{1, 4} \and
Kai Han\inst{1, 5} \and
Vineeth N Balasubramanian\inst{2}}
\authorrunning{Joseph \etal}
\institute{Google Research \and
Indian Institute of Technology Hyderabad \and
Indian Institute of Science, Bangalore \and
Indian Institute of Technology Kanpur \and
The University of Hong Kong \\
\email{\{cs17m18p100001,~vineethnb\}@iith.ac.in, somabiswas@iisc.ac.in,} \\
{\tt\small \{sujoyp,~gauravaggarwal\}@google.com, piyush@cse.iitk.ac.in, kaihanx@hku.hk}}
% \end{comment}
%******************
\maketitle

\begin{abstract}
Humans possess an innate ability to identify and differentiate instances that they are not familiar with, by leveraging and adapting the knowledge that they have acquired so far. Importantly, they achieve this without deteriorating the performance on their earlier learning. Inspired by this,
% in this paper, 
we identify and formulate a new, pragmatic problem setting of \textit{\gcdwfshort:~\gcdwf}, which tasks a machine learning model to incrementally discover novel categories of instances from unlabeled data, while maintaining its performance on the previously seen categories.
% that was used to bootstrap the model.
% , without requiring any task identifying information. 
We propose 1) a method to generate pseudo-latent representations which act as a proxy for (no longer available) labeled data, thereby alleviating forgetting, 2) a mutual-information based regularizer which enhances unsupervised discovery of novel classes, and 3) a simple \LII which aids generalized inference when the testing data contains instances form both seen and unseen categories. 
We introduce experimental protocols based on CIFAR-10, CIFAR-100 and ImageNet-1000 to measure the trade-off between knowledge retention and novel class discovery. Our extensive evaluations reveal that existing models catastrophically forget previously seen categories while identifying novel categories, while our  method is able to effectively balance between the competing objectives. 
% We operate in a generalized setting, where task identifying information is absent during inference. 
We hope our work will attract further research into this newly identified pragmatic problem setting. 

% \vspace{-10pt}
\end{abstract}

% \vspace{-25pt}
\section{Introduction}
% \vspace{-10pt}

Over the last decade, deep learning algorithms have achieved remarkable performances on multiple computer vision tasks \cite{duan2019centernet,mohan2021efficientps,tolstikhin2021mlp,bulat2020toward,sauer2021projected}, even outperforming humans on many of them. These algorithms are specialised to work well in their strictly designed problem setting, but are brittle when the assumptions are relaxed. We closely analyse one such setting here. Current image classification models assume availability of training examples of all classes of interest. Once trained and deployed, it recognises instances of classes that it has been taught. An instance outside this set of classes may be wrongly classified into one of the known classes often with high confidence \cite{tang2021codes,zhou2021learning,scheirer2012toward,geng2020recent}. In contrast, humans can easily identify instances that they do not know, and even differentiate among them. 
%%%%%%%%% FIGURE STARTS
\begin{figure}[t]
\includegraphics[width=\columnwidth]{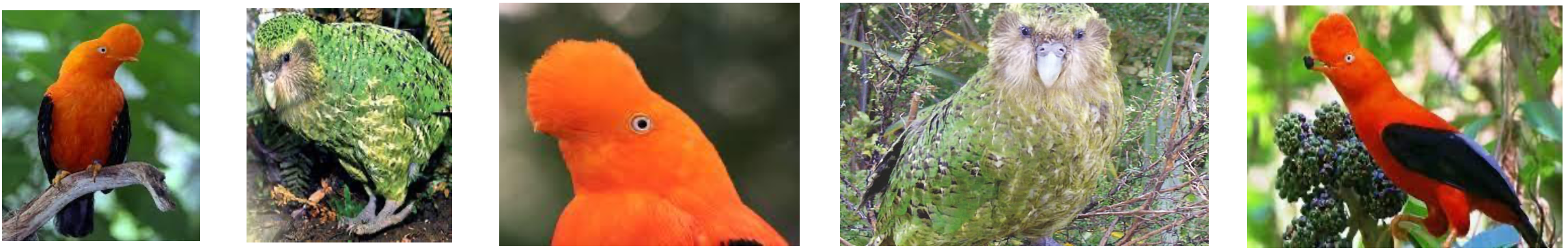}
% \vspace{-15pt}
\caption{
% \underline{Thought Experiment}: Let us first do a simple exercise. Please take a moment to analyse these images before reading further. Done? Thank you very much. We motivate our problem setting based on your intuitive thoughts in the following paragraph. 
%  discuss what we think about these in the first paragraph that follows
Our existing knowledge about birds helps us to easily identify two groups in these images even if we have not seen images of these bird species before. At the same time, unsupervisedly discovering these novel categories does not make us forget about previously seen categories. Motivated by this observation, we propose \textit{\gcdwfshort} setting and a methodology to instill this capability into machines.
% , though it is not obvious to identify these as \cls{Tunki} and \cls{Kakapo}. In our work, we formalise this setting as \textit{\gcdwf}, and propose a methodology for the same. 
}
\label{fig:teaser}
% \vspace{-20pt}
\end{figure}
%%%%%%%%% FIGURE ENDS
To aid our discussion, let us glance through the set of 
% (assumed previously unseen) 
images in \Cref{fig:teaser}. We naturally concur the following: ``These birds are not like anything that we have seen before, but these images do seem to belong to two distinct categories". Importantly, we are able to do this grouping without having access to training images from other objects that we have learnt during our lifetime. Secondly, the ability to do this grouping does not impede us from identifying other kinds of birds that we are already familiar with. Lastly, we achieve this without explicit information that these instances are from novel categories. Motivated by this intrinsic ability of humans, we propose a problem setting, which we refer to as \textit{\gcdwfshort:~\gcdwf}. 

\begin{wrapfigure}[12]{r}{0.49\textwidth}
\vspace{-22pt}
\includegraphics[width=0.48\columnwidth]{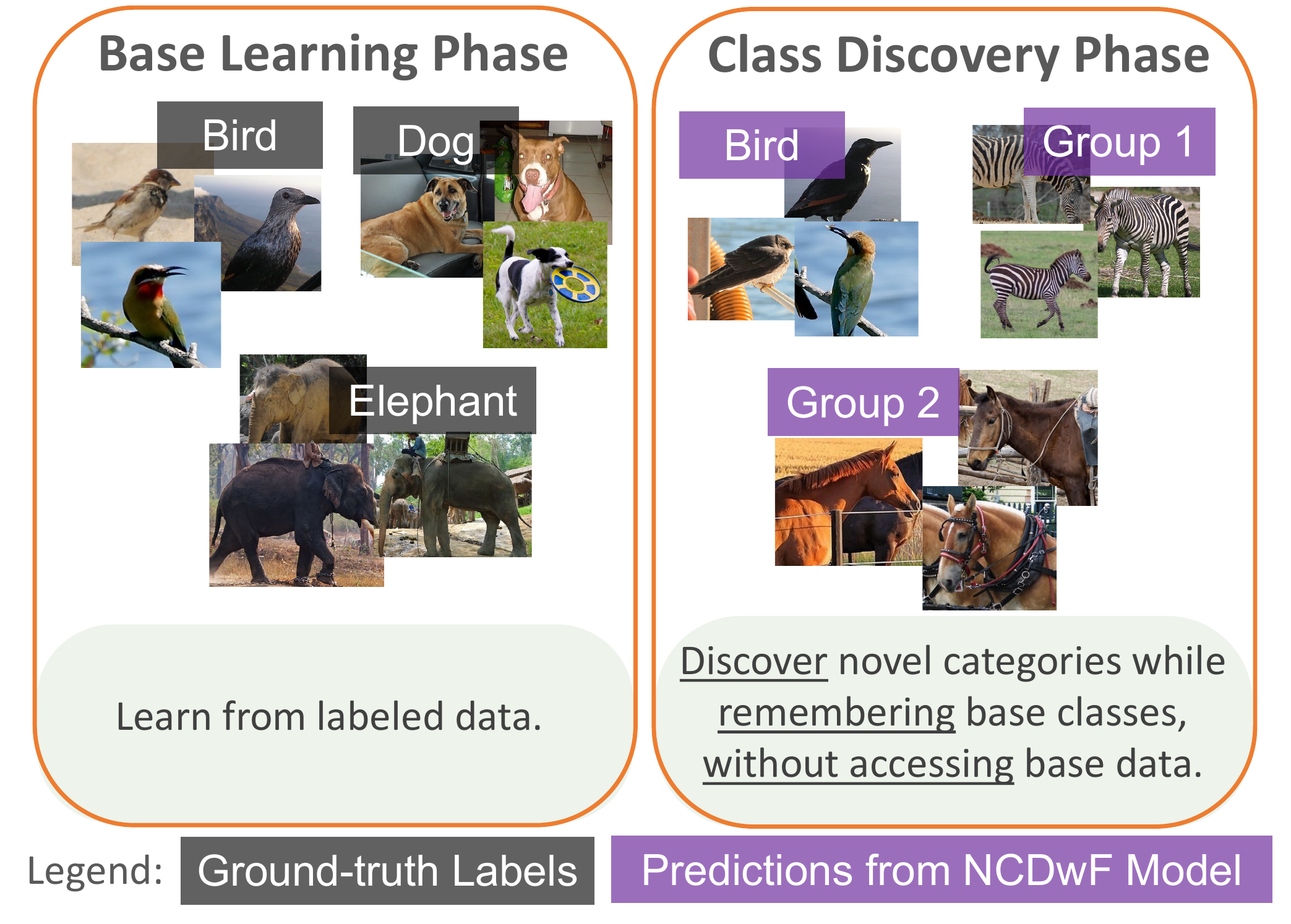}
\vspace{-10pt}
\caption{Summary of  \gcdwfshort setting.
}
\label{fig:setting}
\vspace{-24pt}
\end{wrapfigure} An \gcdwfshort model learns in phases. In the first phase, the model is supervised to learn a few set of classes. In the subsequent phases, the model should automatically identify instances of novel categories from an unlabeled pool containing instances from a disjoint set of classes. While doing so, model \textit{does not have} access to labeled data from the first phase. At any point in time, the model should classify a test instance to one of the labeled or unlabeled classes, without any task identifying information. Here ``task'' refers to whether the test instance belongs to a (known) labeled class or a (novel) unlabeled class. We illustrate the problem setting in \Cref{fig:setting}. After learning about \cls{Bird}, \cls{Dog} and \cls{Elephant} in the first phase, a \gcdwfshort model identifies instances from previously known classes (eg. \cls{Bird}), along with grouping instances of novel categories.  

The \gcdwfshort setting has wide practical applicability: 
% We highlight two of them here: 
1) Consider the recognition component of a robot operating in an open-world. It can be trained in-house with annotated data. Once deployed, it would be of immense value if it can automatically group unknown instances into different groups, along with consistently identifying instances that it has been trained with. 2) Equally interesting would be an online fraud detection system. It can also be trained with a set of known fraud patterns, but it would be hard to speculate emerging frauds. An incremental class discovery model can not only identify novel frauds, but also group them separately, alongside identifying known fraud types, adding immense practical utility. Labeled data that was used to train both these models in-house cannot be accessed while identifying novel instances due to storage and privacy concerns.

NCDwF is closely related to Novel Class Discovery (NCD)\cite{han2019automatically} but it extends NCD in several key aspects.
% First, in NCD, unlabeled instances are assumed to come from only novel classes. 
First, existing NCD methods assume access to both labeled and unlabeled data at training time, which is unlikely to hold for many real applications. 
Second, at test time, current NCD methods assume access to the ``task'' information, i.e., the information whether an unlabeled instance is from a labeled class or not. 
% Novel Class Discovery (NCD) setting \cite{han2019automatically} is probably the closest to our proposed \gcdwf setting. In NCD, labeled instances from a disjoint set of classes can be used to learn a model which can group instances from an unlabeled pool. Interestingly, the performance on the labeled classes is not always accounted for in NCD. Further, most of these methods \cite{han2019automatically,Fini2021unified,zhong2021neighborhood,zhao2021novel} assume access to entire labeled data while training the model to identify instances in unlabeled pool. 
In \gcdwfshort, we relax these assumptions to propose a more pragmatic extension to NCD setting, mirroring real world demands.

Our methodology subtly makes use of the classifier trained on the labeled data to reduce forgetting and improve class discovery. To make up for the lack of labeled examples from previous classes during the unsupervised novel class discovery phase, we identify regions in the latent space by ``inverting" the classifier's discriminative information. Additionally, we ensure that these inverted pseudo-latent representations are close to the true class representations as explained in \Cref{sec:alleviating_forgetting}. These class specific pseudo-representations can be replayed along with unlabeled data to address forgetting. We note that this method is cheaper than the generative modelling alternatives, and does not require any labeled image to be stored and replayed. 
% The closed-set classifier makes over-confident predictions on the unlabeled data. 
In \Cref{sec:enhancing_class_discovery}, we show that maximizing the mutual information between the labeled logits and the unlabeled logits acts as an effective regularizer to enhance class discovery.
% In a practical setting, an inference data can belong to any of the classes of interest. 
The proposed setting calls for a generalized, task-agnostic inference where a test instance may belong to labeled or the unlabeled classes, and such identifying information would be absent during inference. We propose to learn a \LII to help us with this discrimination in \Cref{sec:task-agnostic-inference}.

\vspace{2pt}
\noindent To summarize, our key contributions are as follows:
\begin{itemize}[leftmargin=*,topsep=0pt, noitemsep]
\item We propose a pragmatic generalization to the NCD setting called \gcdwf (NCDwF).
% \item We extend the NCD setting introduce a pragmatic extension to NCD setting called \gcdwf.
\item We introduce an effective method which unsupervisedly discovers novel classes, while retaining performance on the labeled classes used to initialize the model.
\item We introduce experimental setting and evaluation protocol for the new setting.
\item When compared with prominent class-discovery methods \cite{Fini2021unified,han2019automatically,zhong2021neighborhood} adapted to our proposed setting, our methodology achieves improved class-discovery performance with significantly less forgetting.
\end{itemize}

% We propose a novel latent inversion based replay strategy to address the inherent forgetting, and introduce a mutual information based regularizer to improve class discovery.
% \vspace{-0.5em}
\section{Related Works}
% \vspace{-1em}
Here, we analyse how \gcdwfshort differs from existing related settings, followed by a survey of research efforts in Incremental Learning and Novel Class Discovery.

% \vspace{-5pt}
\subsection{Relation with Existing Settings} \label{sec:relationWithSettings}
\begin{table}[t]
% \vspace{-10pt}
\centering
\caption{We summarise related problem settings here.
% The table shows how \gcdwfshort is related to existing related problems. 
We note that \gcdwf is most pragmatic when compared with the others.
% SSL, DA and TL refers to Semi-Supervised Learning, Domain Adaptation and Transfer Learning approaches. 
\checkmark, $\times$ and $-$ indicates \cls{yes}, \cls{no}, and \cls{not-applicable} respectively. More discussion in \Cref{sec:relationWithSettings}.
% \vspace{-5pt}
}
\label{tab:comparison}
\resizebox{\textwidth}{!}{%
\begin{tabular}{>{\kern-\tabcolsep}l|cccc<{\kern-\tabcolsep}}
\toprule
\rowcolor{Gray} \multicolumn{1}{c|}{Characteristics ($\rightarrow$)} & \multicolumn{4}{c}{Data from a future step:} \\ \midrule
\rowcolor{Gray} \multicolumn{1}{c|}{Settings ($\downarrow$)} & \begin{tabular}[c]{@{}c@{}}can contain disjoint \\ set of classes.\end{tabular} & \begin{tabular}[c]{@{}c@{}}need not have \\ side information.\end{tabular} & \begin{tabular}[c]{@{}c@{}}can make use of \\ a model bootstrapped \\ with labeled data.\end{tabular} & \begin{tabular}[c]{@{}c@{}}can be fully \\ unlabeled.\end{tabular} \\ \midrule
Semi-supervised Learning & $\times$ & $-$ & \checkmark & $-$ \\
Zero-shot learning & \checkmark & $\times$ & \checkmark & \checkmark \\
One / Few-shot learning & \checkmark & $\times$ & \checkmark & $\times$ \\
Clustering & $-$ & $-$ & $\times$ & \checkmark \\
Incremental Learning & $-$ & \checkmark & $-$ & $\times$ \\
\rowcolor{highlight} \gcdwfshort & \checkmark & \checkmark & \checkmark & \checkmark \\ \bottomrule
\end{tabular}%
}
% \vspace{-10pt}
\end{table}

We systematically analyse how our proposed setting is related to research efforts in related problem spaces in \Cref{tab:comparison}. \gcdwfshort methods incrementally discover novel category of instances from an unlabeled pool by utilizing the knowledge from a disjoint set of labeled instances. At inference stage, the model should be consistent in classifying instances to any of labeled or unlabeled classes, without any task identifying information. In semi-supervised learning approaches \cite{chapelle2009semi,van2020survey}, the labeled and unlabeled data comes from the same set of classes. Zero-shot learning methods \cite{xian2017zero,romera2015embarrassingly} require prior knowledge of extra semantic attribute information about the unlabeled classes. Few-shot learning methods \cite{allen2019infinite,zhang2021self,snell2017prototypical} additionally require a few of the unlabeled instances to be labeled. Similar instances are grouped together by clustering algorithms \cite{xu2005survey,du2010clustering}, but they cannot make use of labeled instances from a disjoint set of classes. Incremental learning methods \cite{rebuffi2017icarl,AGEM,kj2020meta} learn a single model across tasks, but data for each incremental task is fully annotated. Methods that perform out-of-distribution detection \cite{liu2020energy,pimentel2014review} and open-set learning \cite{scheirer2012toward,geng2020recent} identify instances significantly different from the training data distribution as novel samples, but do not identify sub-groups within these identified instances automatically. To the best of our knowledge, the proposed setting has minimal assumptions and is most pragmatic, when compared to these settings.

% \vspace{-10pt}
\subsection{Incremental Learning} The core focus of incremental learning methods is to alleviate the catastrophic forgetting of neural networks \cite{french1999catastrophic,mccloskey1989catastrophic}, when learning a single model across a sequence of tasks. Regularization based methods \cite{li2017learning,rebuffi2017icarl,castro2018end,wu2019large,douillard2020podnet,liu2020mnemonics} ensure that the parameter adaptations for the new task will be optimal for all the tasks learned so far. Another kind of approach either stores or generates exemplar images for all the tasks introduced to the model so far and replays them while learning a new task \cite{rebuffi2017icarl,liu2020mnemonics,belouadah2019il2m,kj2020meta}. This ensures consistency across all tasks. Dynamically expanding and parameter isolation methods \cite{rusu2016progressive,rajasegaran2019adaptive,rajasegaran2019random,abati2020conditional,liu2021adaptive} form a third class of methods to address forgetting. 
All these methods require access to labeled instances for all the tasks. In contrast, \gcdwf models identify novel categories from unlabeled data which the model encounters incrementally - without forgetting how to identify instances in the labeled classes which were initially used to bootstrap the model.

% \vspace{-10pt}
\subsection{Novel Class Discovery}
Earlier methods like MCL \cite{Hsu19_MCL} and KCL \cite{Hsu18_L2C} for general transfer learning across domains and tasks  meta-learn a binary similarity function from labeled data and use it to discover classes in unlabeled data. DTC \cite{han2019learning} formalized the problem of Novel Class Discovery and introduced a method based on Deep Embedded Clustering \cite{xie2016unsupervised} for NCD, by pre-training it on the labeled data followed by learning-based clustering. RS \cite{han2019automatically} first pretrains the model on the labeled and the unlabeled data with self-supervision and uses ranking statistics to generate pseudo-labels for learning the novel categories. This has been further extended by Zhao and Han \cite{zhao2021novel} to further take local spatial information into account. NCL \cite{zhong2021neighborhood} introduces contrastive learning and OpenMix \cite{zhong2021openmix} uses a convex combination of labeled and unlabeled instances to enhance class discovery. UNO \cite{Fini2021unified} learns a unified classifier which identifies labeled and unlabeled instances using ground-truth labels and pseudo-labels respectively. 
% They also introduce a task-agnostic evaluation protocol. 
Joseph \etal~\cite{joseph2022spacing} uses cues from multi-dimensional scaling to enforce latent space separability, while Jia \etal~\cite{jia21joint} leverages contrastive learning with WTA hashing to enhance class discovery.
% to discover new categories in videos and images.  

% RS \cite{han2019automatically}, NCL \cite{zhong2021neighborhood}, OpenMix \cite{zhong2021openmix} and UNO \cite{Fini2021unified}
Existing methods for NCD require access to labeled and unlabeled instances together to discover novel categories, which limits their practical applicability. Most of these methods also assume the unlabeled data only contains instances from new classes or assume the information that whether an unlabeled instance is from new classes is known.
% Most of these methods measure the competency only on the class discovery, where the performance on the labeled data can be sacrificed, which is also limiting. 
The concurrent work by Vaze \etal~\cite{vaze22generalized} extends NCD to a generalized setting where the unlabeled instances may come from both old and new classes, while still requiring access to labeled and unlabeled instances jointly.
In contrast, with \gcdwf, we introduce a staged learning and account for the performance on both labeled and the unlabeled data, without requiring access to the labeled data when learning on unlabeled data to discover new classes. Meanwhile, at test time, we do not assume the unlabeled images are only from new classes nor require to know whether an unlabeled image is from a new class or an old one.
% Storing labeled data and reusing them can incur storage expense and may violate privacy policies.

\section{\gcdwf} \label{sec:methodology}
We formally define \gcdwf in \Cref{sec:formulation}. \gcdwfshort models should balance between two competing goals: alleviating forgetting of labeled classes without impairing unsupervised novel class discovery capability. \Cref{sec:alleviating_forgetting} and \Cref{sec:enhancing_class_discovery} explain how we achieve these objectives. In \Cref{sec:task-agnostic-inference}, we propose \LII, which helps with task-agnostic inference.

% \vspace{-5pt}
\subsection{Formulation}
\label{sec:formulation}
Given a labeled data pool $D_{lab}=\{(\bm{x}_i, y_i) \sim P(\mathcal{X}, \mathcal{Y}_{lab})\}$, \gcdwf  aims to learn a model $\bm{\mathrm{\Psi}}$ that would identify novel category of instances from an unlabeled data pool $D_{unlab}=\{(\bm{x}_i) \sim P(\mathcal{X} \mid \mathcal{Y}_{unlab})\}$, along with recognizing instances from $D_{lab}$.
% without degrading its performance on $D_{lab}$.
The label space of $D_{lab}$ and $D_{unlab}$ are disjoint, i.e., $\mathcal{Y}_{lab} \cap \mathcal{Y}_{unlab} = \emptyset$. Further, while discovering novel categories, $D_{lab}$ cannot be accessed.
The problem setting naturally induces a multi-stage learning where $\bm{\mathrm{\Psi}}$ initially learns a representation to identify instances in $D_{lab}$, which would then be re-purposed to identify novel instances unsupervisedly.
The main challenge involved in learning such a $\bm{\mathrm{\Psi}}$ is to accurately group instances from $D_{unlab}$ into semantically meaningful categories, without degrading its performance on identifying the labeled instances from $D_{lab}$. Additionally, such a segregation should be done in a generalized fashion, where task identifying information would be absent during inference. 

%%%%%%%%% FIGURE STARTS
\begin{figure}[t]
\includegraphics[width=\columnwidth]{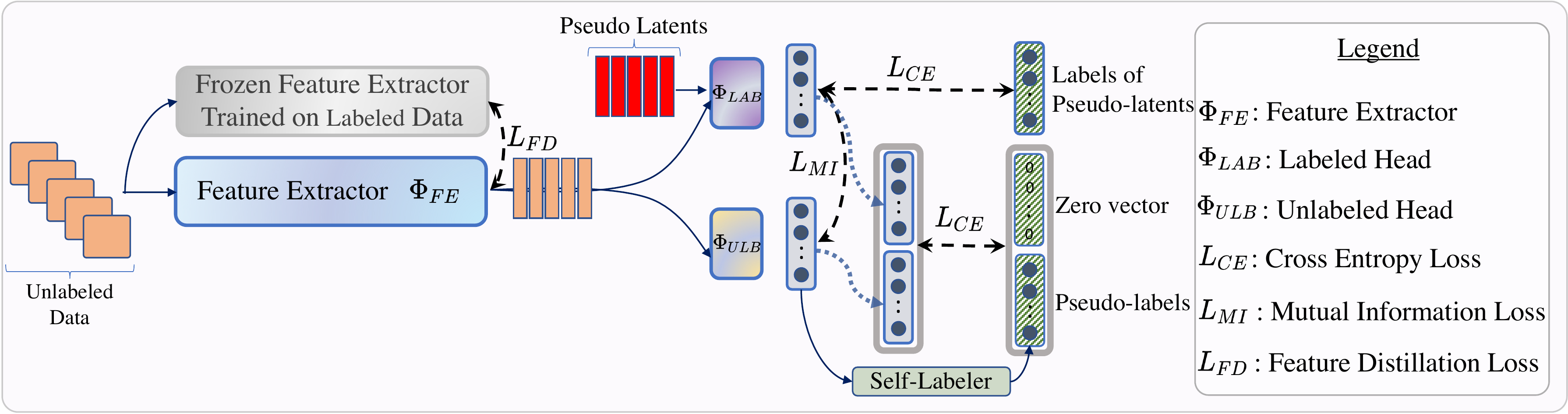}
% \vspace{-15pt}
\caption{
The figure illustrates how our proposed approach discovers novel categories, while retaining its performance on the labeled data.
% The main components of the proposed approach while it incrementally discovers novel category of instances from unlabeled data is illustrated here. 
The network consists of a feature extractor $\bm{\mathrm{\Phi}}_{FE}$ shared between the labeled head $\bm{\mathrm{\Phi}}_{LAB}$ and unlabeled head $\bm{\mathrm{\Phi}}_{ULB}$. We generate pseudo-latents and replay them through the labeled head to reduce its forgetting (\Cref{sec:alleviating_forgetting}), and guide the unlabeled head learning through the pseudo-labels and the mutual-information based regularizer (\Cref{sec:enhancing_class_discovery}). 
% The model trained only on labeled data, shown in gray, provides additional regularization while learning the model end-to-end. 
% (\textit{best viewed in color}).
}
\label{fig:architecture}
% \vspace{-10pt}
\end{figure}
%%%%%%%%% FIGURE ENDS

We illustrate the main components of our architecture that help to discover novel categories without forgetting labeled instances in \Cref{fig:architecture}.
Without loss of generality, we assume that the model $\bm{\mathrm{\Psi}}$ consists of a feature extractor $\bm{\mathrm{\Phi}}_{FE}$, one head for classifying the labeled instances $\bm{\mathrm{\Phi}}_{LAB}$, and another head for discovering novel categories $\bm{\mathrm{\Phi}}_{ULB}$. 
The feature extractor is shared between both heads. 
% While discovering novel categories, we assume access to the model trained only on labeled classes (shown in gray).  
Pseudo-latents (shown in red) serve as a proxy for labeled data during category discovery. 
Pseudo-labels from the self-labeler and the  regularization enforced by the mutual-information loss guide the learning of unlabeled head. 
% We explain this in detail in \Cref{sec:enhancing_class_discovery}. 
A frozen model trained only on labeled classes
(shown in gray) 
is also used to regularise the model via feature-distillation loss $L_{FD}$ \cite{hinton2015distilling}.
We apply an L2 loss between backbone features from the model trained on labeled data $\bm{\mathrm{\Phi}}_{FE}^{lab}(\bm{x})$ and current model $\bm{\mathrm{\Phi}}_{FE}^{~} (\bm{x})$ as follows:
$L_{FD} = \lVert \bm{\mathrm{\Phi}}_{FE}^{~}(\bm{x}) - \bm{\mathrm{\Phi}}_{FE}^{lab}(\bm{x}) \rVert_2 $. Such feature distillation loss has been used in incremental detectors \cite{kj2021incremental} and is simpler than the Less-Forget constraint from LUCIR \cite{hou2019learning}.
The whole model is learned end-to-end, where the feature extractor is free to adapt itself to improve class-discovery, while maintaining its performance of recognizing instances from labeled classes. 
% In the following sections, we explain the methodological components that enhances novel class discovery without forgetting. 

% \vspace{-5pt}
\subsection{Retaining Performance on Labeled Classes}\label{sec:alleviating_forgetting}

It would be of immense practical value if a model that is trained in-house with labeled data is able to identify novel category of instances, when deployed in an open world. When the network $\bm{\mathrm{\Psi}}$ improves its ability to group instances of novel categories from $D_{unlab}$, it may drastically fail to retain its performance on recognizing the labeled instances, which were learned from $D_{lab}$, like the well-known \emph{catastrophic forgetting} in lifelong learning \cite{french1999catastrophic,mccloskey1989catastrophic}. This happens as the model cannot be jointly optimised for category discovery and classification of the known instances due to the unavailability of $D_{lab}$. 
% Exemplar replay is a simple yet powerful technique to address the inherent forgetting, at the cost of storing few data instances. 

\begin{figure}[t]
\includegraphics[width=\columnwidth]{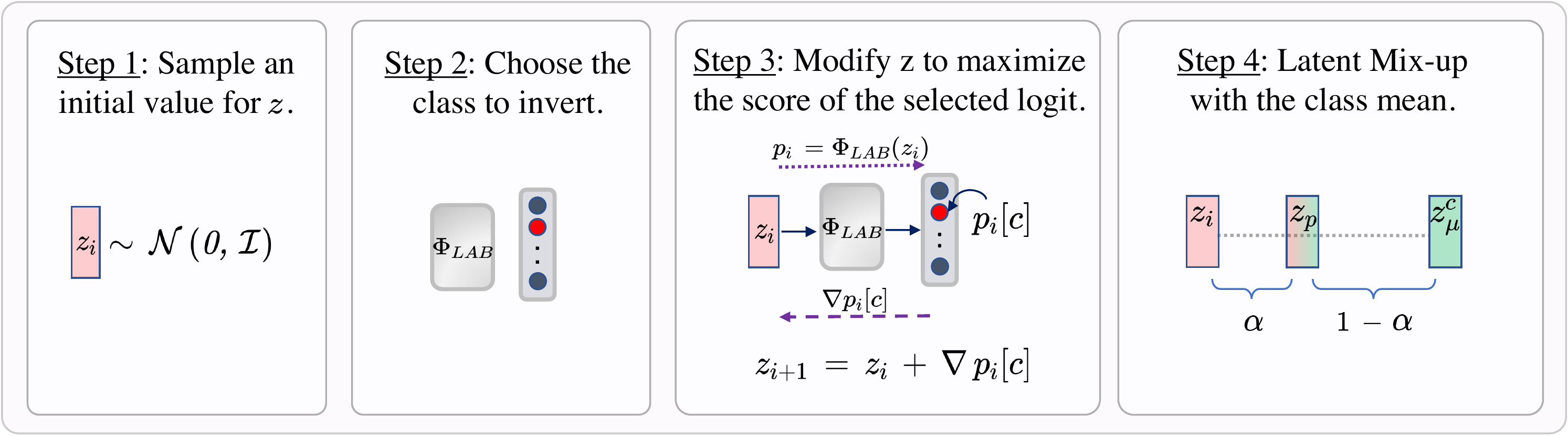}
% \vspace{-15pt}
\caption{In the \gcdwfshort setting, labeled data cannot be accessed while discovering novel categories. We propose to generate pseudo-latent representations, which can act as effective proxy for the labeled data representations, by inverting the discriminative information from the trained labeled head as shown in Step 1, 2 and 3. Step 4 helps to induce extra semantic information into these synthesised pseudo-latents. 
% (\textit{best viewed in color}).
}
\label{fig:generating_pseudo_latents}
% \vspace{-10pt}
\end{figure}

We propose a novel methodology that would generate pseudo latent representations, which can act as a proxy for the latent representations of the labeled training data. 
We make use of the classifier $\bm{\mathrm{\Phi}}_{LAB}$ that was trained solely on the labeled classes to generate these pseudo-latent representations $\bm{z}_p$. We explicitly learn these such that it maximally activates a selected class of interest.
% , from the vector of predicted logits $\bm{p}$.
\Cref{fig:generating_pseudo_latents} summarizes the steps involved to invert the latent knowledge from the classification head. First, we sample $\bm{z}_i$ from a standard Normal distribution, then we select the specific class $c$ for which we would like to generate the pseudo-latents. Next, we do a gradient ascent on $\bm{z}_i$ such that the score for the selected class $c$ would be higher for the predicted logit vector $\bm{p}_i = \bm{\mathrm{\Phi}}_{LAB}(\bm{z}_{i})$. Importantly, the parameter of $\bm{\mathrm{\Phi}}_{LAB}$ are frozen, while carrying out the latent inversion $\bm{z}_{i+1} = \bm{z}_{i} + \nabla \bm{p}_{i}[c]$, where $\bm{p}_{i} = \bm{\mathrm{\Phi}}_{LAB}(\bm{z}_{i})$. Next, we do \textit{mixup} \cite{zhang2018mixup} in latent space between inversed latent $\bm{z}_i$ and corresponding class mean of labeled training instances $\bm{z}^{c}_{\mu}$. 
%%%%%%%%% FIGURE STARTS
%%%%%%%%% FIGURE ENDS
\Cref{algo:replay} summarises the steps to generate the latent pseudo-dataset $D_{pseudo}$. In Lines 4 - 7, we invert the latents of the specific class $c$, using $\bm{\mathrm{\Phi}}_{LAB}$. In Lines 8 and 9, we first select a mixing coefficient $\alpha$, and do a linear combination of the inverted latent $\bm{z}_{L}$ and the class mean $\bm{z}^{c}_{\mu}$.   The class means $\bm{K}$ can be computed and stored after the first phase of learning labeled instances. This mixing operation helps to smoothen the latent space and impart additional semantic information to $\bm{z}_p$.
The labeled pseudo-dataset $D_{pseudo}$ is replayed while learning to identify novel categories, to arrest forgetting in labeled head of $\bm{\mathrm{\Psi}}$. 
% The latent representations from the backbone network are also regularized by distilling from the backbone that is trained only on labeled data as shown in \Cref{fig:architecture}.  

% \vspace{-10pt}
\begin{algorithm}
\small
\caption{Algorithm \textsc{GeneratePseudoDataset}}
\label{algo:replay}
\begin{algorithmic}[1]
\Require{Number of labeled classes: $M$; Number of pseudo-data per class: $E$; Number of inversion iterations: $L$; labeled head: $\bm{\mathrm{\Phi}}_{LAB}$; Class Means: $\bm{K} = \{\bm{z}^{1}_{\mu}, \cdots, \bm{z}^{M}_{\mu}\}$; Parameters of the Beta-distribution: $\gamma$, $\rho$.
}
\Ensure{Labeled pseudo-dataset: $D_{pseudo}$.
}
\State $D_{pseudo} \leftarrow [~]$
\For{c in (1 $\cdots$ M)}
\For{e in (1 $\cdots$ E)}
\State $\bm{z}_1 \sim \mathcal{N}(0, \bm{I})$
\For{i in (1 $\cdots$ L)} \Comment{Latent Inversion.}
\State $\bm{p}_{i} = \bm{\mathrm{\Phi}}_{LAB}(\bm{z}_i)$
\State $\bm{z}_{i+1} = \bm{z}_{i} + \nabla \bm{p}_{i}[c]$
\EndFor 
\State $\alpha \leftarrow Beta(\gamma, \rho)$
\State $\bm{z}_{p} = \alpha \bm{z}_{L} + (1 - \alpha) \bm{z}^{c}_{\mu}$ \Comment{Latent \textit{mixup} \cite{zhang2018mixup} with class mean.}
\State $D_{pseudo} \leftarrow D_{pseudo} + (\bm{z}_{p}, c)$
\EndFor
\EndFor
\State \Return $D_{pseudo}$
\end{algorithmic}
\end{algorithm}

% \vspace{-10pt}
\subsection{Enhancing Class Discovery}\label{sec:enhancing_class_discovery}
Motivated by the success of self-labelling algorithms in self-supervised learning \cite{asano2020self,caron2020unsupervised}, Fini \etal~\cite{Fini2021unified} re-purposes it to automatically generate pseudo labels for the unlabeled data. These labels are used to train the unlabeled head $\bm{\mathrm{\Phi}}_{ULB}$. A key characteristic of such a self-labelling function would be to discourage degenerate solutions. This is explicitly enforced by  pseudo-labeling a mini-batch such that the data-points are split uniformly across all the $N$ classes in the unlabeled pool 
% it is partitioned into equally sized subset
\cite{caron2020unsupervised,Fini2021unified}. Formally, let $\bm{P} = \{\bm{p}_1, \bm{p}_2, \cdots, \bm{p}_{B}\}$ be the predictions from $\bm{\mathrm{\Phi}}_{ULB}$ for a mini-batch of unlabeled data. Let each mini-batch contains $B$ instances. We seek to find label assignment $\bm{Q}^* = \{\bm{q}_1, \bm{q}_2, \cdots, \bm{q}_{B}\}$, such that it respects heterogeneous cluster assignment. This setting can be reduced to solving the following optimal transport problem \cite{asano2020self,caron2020unsupervised}: 
\begin{equation}
    \bm{Q}^* = \max_{\bm{Q} \in \bm{\mathcal{Q}}} \Tr (\bm{Q}^\top\bm{P}) - \sum_{i,j} \bm{Q}_{ij} \log \bm{Q}_{ij}
    \label{eq:pseudo_label}
\end{equation}
where $\bm{\mathcal{Q}}$ is the transportation polytope defined as $\bm{\mathcal{Q}} = \{\bm{Q} \in \mathbb{R}_{+}^{N \times B} | \bm{Q}\bm{1}_{\bm{B}} = \frac{1}{N}\bm{1}_{N}, \bm{Q}^\top\bm{1}_{{N}} = \frac{1}{B}\bm{1}_{B}\}$. An iterative Sinkhorn-Knopp algorithm \cite{cuturi2013sinkhorn} can be used to solve Eqn.~\ref{eq:pseudo_label} to find the optimal pseudo-label $\bm{Q}^*$. 
% Thereafter, the parameters of the unlabeled head and the backbone network is updated to minimize the standard cross-entropy loss between $\bm{P}$ and $\bm{Q}^*$.
% The assumption that each mini-batch will have data from all classes in unlabeled set is fallible. 
% We propose a method to enhance the semantic consistency of such pseudo labels next. 
% We propose a method to induce the semantic consistency while learning the unlabeled head.
The assumption that each mini-batch will be partitioned into all unlabeled classes is fallible. This would lead to noisy pseudo-labels that are not semantically grounded.
We are motivated by the observation that labeled head confidently predicts unlabeled data-points into one of the semantically related known categories. For instance, a \cls{motorcycle} gets misclassified into semantically related \cls{bicycle}, and not into other classes that are completely unrelated (more examples in \Cref{sec:misclassification}).
We propose a method which complements the learning via the self-labeled pseudo label  by using the semantic information that is available for free within the labeled head.
% Self-labeling algorithms strictly partitions each mini-batch into all the unlabeled classes, 
% When properly used, this semantic information can guide the leaning of the unlabeled head 

In the first stage of \gcdwfshort, instances from the labeled data pool $D_{lab}$ would be introduced to the model $\bm{\mathrm{\Psi}}$. We train the feature extractor $\bm{\mathrm{\Phi}}_{FE}$, and the labeled head $\bm{\mathrm{\Phi}}_{LAB}$ with $D_{lab}$. When we pass an instance from $D_{unlab}$ through $\bm{\mathrm{\Phi}}_{LAB} \circ \bm{\mathrm{\Phi}}_{FE} (\bm{x})$, the unlabeled instances would be predicted to one of the labeled classes consistently. 
% This overconfident nature of deep networks is extensively studied in open-set literature \cite{tang2021codes,zhou2021learning,scheirer2012toward,geng2020recent}. 
We make use of these overconfident predictions from the labeled head to guide unknown identification in $\bm{\mathrm{\mathrm{\Phi}}}_{ULB}$. An information theoretic approach to achieve this would be to maximize the mutual information between the predictions from labeled head and unlabeled head, such that we can transfer semantic information from the labeled to unlabeled head, as motivated by~\cite{ahn2019variational}. Concretely, for an image $\bm{x} \in D_{unlab}$, let $\bm{l} = \bm{\mathrm{\Phi}}_{LAB} \circ \bm{\mathrm{\Phi}}_{FE} (\bm{x})$ denote the logits from the labeled head and $\bm{u} = \bm{\mathrm{\Phi}}_{ULB} \circ \bm{\mathrm{\Phi}}_{FE} (\bm{x})$ denote the logits from the unlabeled head. $\bm{l}$ and $\bm{u}$ can be of different dimensions: $\bm{l} \in \mathbb{R}^M$ and $\bm{u} \in \mathbb{R}^N$. We intend to guide the learning of $\bm{\mathrm{\Phi}}_{ULB}$ by maximizing the mutual information ${I}(\bm{l};\bm{u})$ between $\bm{l}$ and $\bm{u}$, which we can expand as follows:

\begin{align}
I(\bm{l};\bm{u}) & = H (\bm{l}) - H(\bm{l}|\bm{u}) \nonumber \\
  & = - \mathbb{E}_{\bm{l}} [\log p(\bm{l})] + \mathbb{E}_{\bm{l}, \bm{u}}[\log p(\bm{l}|\bm{u})]
\label{eq:MutualInformation}
\end{align}

\noindent where $H (\bm{l})$ refers to the entropy of $\bm{l}$ and $H(\bm{l}|\bm{u})$ is the conditional entropy between the random variables $\bm{l}$ and $\bm{u}$, sampled from a probability distribution $p(.)$. Numerically computing exact mutual information is intractable, and hence we resort to a variational approximation $q(\bm{l}|\bm{u})$ to true distribution $p(\bm{l}|\bm{u})$ \cite{agakov2004algorithm,ahn2019variational} as follows:

\begin{align}
I(\bm{l};\bm{u}) 
  & = - \mathbb{E}_{\bm{l}} [\log p(\bm{l})] + \mathbb{E}_{\bm{l}, \bm{u}}[\log p(\bm{l}|\bm{u})]\nonumber \\
  & \approx - \mathbb{E}_{\bm{l}} [\log p(\bm{l})] + \mathbb{E}_{\bm{l}, \bm{u}}[\log q(\bm{l}|\bm{u})]+ \mathbb{E}_{ \bm{u}}[KL( p(\bm{l}|\bm{u}) ~||~ q(\bm{l}|\bm{u}))]\nonumber \\
  & \geq - \mathbb{E}_{\bm{l}} [\log p(\bm{l})] + \mathbb{E}_{\bm{l}, \bm{u}}[\log q(\bm{l}|\bm{u})]
\label{eq:VariationalDistribution}
\end{align}

We assume the variational distribution to be a Gaussian, with a learnable mean function $\bm{\mu}_{\bm{\theta}}(\bm{u})$ and variance function $\bm{\sigma}_{\bm{\omega}}$.
% , where $\bm{u}$ are the logits from the unlabeled head. 
This would extend the derivation in Eqn.~\ref{eq:VariationalDistribution} to the following:
\begin{align}
I(\bm{l};\bm{u}) 
  & \geq - \mathbb{E}_{\bm{l}} [\log p(\bm{l})] + \mathbb{E}_{\bm{l}, \bm{u}}[\log q(\bm{l}|\bm{u})] \nonumber \\
  & = - \mathbb{E}_{\bm{l}} [\log p(\bm{l})] + \mathbb{E}_{\bm{l}, \bm{u}}[\sum_{i = 1}^{M}\log {\sigma}_{\omega}^{i} + \frac{(l^i - \mu_{\theta}(\bm{u}))^2}{2(\sigma_{\omega}^{i})^2}]
\label{eq:gaussian}
\end{align}

The parameters $\bm{\theta}$ and $\bm{\omega}$ of the mean and variance functions, the unlabeled head $\bm{\mathrm{\Phi}}_{ULB}$ and the feature extractor $\bm{\mathrm{\Phi}}_{FE}$ would be updated to maximize the mutual information between $\bm{l}$ and $\bm{u}$. 
% As the parameters of the labeled head is not optimized, 
As the first term in the RHS of Eqn.~\ref{eq:gaussian} is a constant, we can rewrite our mutual information based loss as below. $L_{MI}$ is minimised along with the standard cross-entropy loss between pseudo labels and the predictions from the unlabeled head $\bm{\mathrm{\Phi}}_{ULB}$.
\begin{equation}
    L_{MI} = - I(\bm{l};\bm{u}) \approx - \mathbb{E}_{\bm{l}, \bm{u}}[\sum_{i = 1}^{M}\log {\sigma}_{\omega}^{i} + \frac{(l^i - \mu_{\theta}(\bm{u}))^2}{2(\sigma_{\omega}^{i})^2}]
    \label{eq:MI_loss}
\end{equation}
\subsection{Towards task-agnostic Inference}\label{sec:task-agnostic-inference}
% Requiring task identifying information to route a data sample to the labeled or unlabeled head during inference would reduce the applicability of our method in real scenarios. We design a simple Labeled Instance Identifier (LII) which automates this decision making. LII is learned as a binary classifier which discriminates between pseudo-latents, which acts as a proxy for labeled data and latent representations of the unlabeled data. This intuitive formulation works very well as evident from the experimental results in \Cref{sec:icd_results}.

So far, we introduced an effective mechanism to address forgetting and an intuitive approach to enhance class discovery. Our basic architecture contains a feature extractor $\bm{\mathrm{\Phi}}_{FE}$ which branches off into the labeled head $\bm{\mathrm{\Phi}}_{LAB}$ and the unlabeled head $\bm{\mathrm{\Phi}}_{ULB}$. During inference, if we know whether a sample indeed belongs to one of the labeled classes or not, we could effectively route it to the corresponding head. But, this would limit the applicability in many realistic scenarios. We circumvent this by learning a function, which we call \textit{\LIIshort:~\LII}, which automates this decision. 

\LIIshort is realised as a two layer neural network $\bm{\mathrm{\Phi}}_{\LIIshort}$ which is trained during the class discovery phase. Hence, labeled instances cannot be accessed to learn this binary function. Instead, we use the methodology explained in \Cref{sec:alleviating_forgetting} to generate $N_p$ pseudo-latents $\bm{z}_p$, which would act as a proxy for the labeled data. Using the $N_u$ unlabeled data that we have access to, we extract their latent representations $\bm{z}_u = \bm{\mathrm{\Phi}}_{FE}(\bm{x})$, where $\bm{x} \sim D_{unlab}$. We create a dataset of latent representations $D_{\LIIshort} = (\mathbb{Z}_{\LIIshort}, \mathbb{Y}_{\LIIshort})$ where $\mathbb{Z}_{\LIIshort} = \{\bm{z}_p^i\}_{i=1}^{N_p}\cup \{\bm{z}_u^i\}_{i=1}^{N_u}$ and $\mathbb{Y}_{\LIIshort} = \{0\}_{i=1}^{N_p}\cup \{1\}_{i=1}^{N_u}$. We learn \LIIshort with the following loss function:

\begin{equation}
    L_{\LIIshort} = \frac{1}{N_p + N_u} \sum_{i = 1}^{N_p + N_u} y_i  \log(\bm{\mathrm{\Phi}}_{\LIIshort}(\bm{z}_i)) + (1- y_i)  \log(1 - (\bm{\mathrm{\Phi}}_{\LIIshort}(\bm{z}_i)))
\end{equation}

This simple formulation learns an effective classifier that differentiates labeled instances from others. We show how the learning of $\bm{\mathrm{\Phi}}_{\LIIshort}$ matures with training in \Cref{sec:trainingLLI}. At inference time, given a latent representation of a test instance $\bm{z}_t$, we compute $\bm{\mathrm{\Phi}}_{\LIIshort}(\bm{z}_t)$ and threshold it using $\tau$, to decide on the prediction. We include a sensitivity analysis on $\tau$ in \Cref{sec:trainingLLI}.
% \begin{equation}
%     y_{pred} = \begin{cases}
%       0, & \text{if}\ \bm{\mathrm{\Phi}}_{\LIIshort}(\bm{z}_t) < \tau \\
%       1, & \text{otherwise}
%     \end{cases}
% \end{equation}

\section{Experiments and Results}\label{sec:expr_and_results}
We define the experimental protocol and evaluate our proposed methodology in this section. 
% \gcdwfshort models balance between improving class discovery on the unlabeled data, while retaining its performance on the labeled data.
We formulate five different data splits across three existing datasets and benchmark against three prominent NCD approaches. We explain these in \Cref{sec:expr_settings} followed by the implementation details in \Cref{sec:impl_details} and results in \Cref{sec:icd_results}.

% \vspace{-5pt}
\subsection{Experimental Setting}\label{sec:expr_settings}
\textbf{Dataset and Splits: } We propose to evaluate \gcdwfshort models on CIFAR-10 \cite{krizhevsky2009cifar}, CIFAR-100 \cite{krizhevsky2009cifar} and ImageNet \cite{russakovsky2015imagenet} datasets. Inspired by the data splits used to evaluate Novel Class Discovery methods \cite{han2019automatically,zhong2021openmix,Fini2021unified}, we derive a labeled set and unlabeled set from these datasets. 
For CIFAR-10, we group the first five classes as labeled and the rest as unlabeled. For CIFAR-100, we propose three different groupings: with the first $80$, $50$ and $20$ classes as labeled and the rest an unlabeled. Lastly, for ImageNet, the first $882$ classes are labeled and $30$ classes from the remaining $118$ classes (referred to as split-A in NCD methods \cite{han2019automatically,zhong2021openmix,Fini2021unified}), are learned incrementally. While learning to discover novel categories, the labeled data cannot be accessed. This is an important difference when compared with the existing NCD setting. 
We evaluate the trained model on the test split of the corresponding datasets. 

\noindent \textbf{Baseline Methods: } We compare our proposed approach with three recent and top performing NCD methods: RS \cite{han2019automatically}, NCL \cite{zhong2021neighborhood} and UNO \cite{Fini2021unified}. To ensure fair comparison with these methods, we retrain these models with code from their official repositories, adapted to our proposed incremental setting.  

\noindent \textbf{Evaluation Metrics: } The performance on the labeled data is measured using the standard accuracy metric. Following the practice in Clustering and NCD approaches \cite{han2019automatically,Fini2021unified,luo2009non,tolic2018nonlinear}, we use clustering accuracy to measure the performance of class discovery on unlabeled data. Denoting $y_i$ to be a prediction that the model gives for $\bm{x}_i \in D_{unlab}$, the clustering accuracy is computed as follows:
\begin{equation}
    \text{Clustering Accuracy} = \max_{p ~\in~ perm (\mathcal{Y}_{unlab})} \frac{1}{N_{u}} \sum_{i=1}^{N_{u}} \mathds{1} \{y_i = p(\hat{y}_i)\}
\end{equation}
where $perm (\mathcal{Y}_{unlab})$ is a set of permutations of the unlabeled classes optimally computed via the Hungarian algorithm \cite{kuhn1955hungarian} and $N_{u}$ refers to the number of instances in $D_{unlab}$. This discounts for the fact that the predicted cluster label might not match the exact ground truth label $\hat{y}_i$.

% \vspace{-6pt}
\subsection{Implementation Details}\label{sec:impl_details}
We use ResNet-18 \cite{he2016identity} backbone for all our experiments. We use SGD with momentum parameter of $0.9$ to train the model on mini-batches of size $512$. We use $200$ epochs for each phase. Following our baseline \cite{Fini2021unified}, we also use multi-head clustering and over-clustering for the class discovery head. We strictly follow Fini \etal~\cite{Fini2021unified} for the design choice of the heads and the hyper-parameters. \LIIshort is modeled as a two layer neural network with 128 neurons each, terminating with a single neuron. For generating the pseudo-data, we sample the mixing coefficient $\alpha$ from $Beta(1, 100)$. In the class discovery phase each mini-batch contains $0.25\%$ of pseudo-data.
The models are evaluated in both task-aware and task-agnostic setting. While doing task-aware inference, we assume that task identifying information 
(whether it belongs to any of the labeled class or not) is available with each test sample. In the more pragmatic task-agnostic setting, we use the proposed \LIIshort to make this decision. For fair evaluation, we use \LIIshort both with our approach and the baseline method \cite{Fini2021unified}. 
After deciding on a specific head (either using the ground-truth or \LIIshort), we take the $argmax$ over the logits to generate the prediction.
RS \cite{han2019automatically} and NCL \cite{zhong2021neighborhood} learn a binary classifier per unlabeled class, while UNO \cite{Fini2021unified} and our method learn a classifier that scores via softmax function. 
% Hence, RS \cite{han2019automatically} and NCL \cite{zhong2021neighborhood} cannot be adapted to the task-agnostic setting, as taking $argmax$ over independent binary classifier predictions would degenerate.
\begin{table}[t]
\centering
% \vspace{-17pt}
\caption{Performance of the model in identifying instances of the labeled categories, along with identifying novel categories (`Lab' and `Unlab' columns respectively), after incrementally learning to discover novel categories is recorded below. We note that our pseudo-data based replay and mutual information based regularization can offer improved class discovery while retaining the performance on the labeled classes, in the task-aware and generalized setting. Please find detailed description in \Cref{sec:icd_results}.}
\label{tab:main-results}
\resizebox{\textwidth}{!}{%
\begin{tabular}{@{}lccccccccccccccc@{}}
\toprule
\rowcolor{Gray} \multicolumn{1}{c|}{Settings ($\rightarrow$)} & \multicolumn{3}{c|}{CIFAR-10-5-5} & \multicolumn{3}{c|}{CIFAR-100-80-20} & \multicolumn{3}{c|}{CIFAR-100-50-50} & \multicolumn{3}{c|}{CIFAR-100-20-80} & \multicolumn{3}{c}{ImageNet-1000-882-30} \\ \midrule
\rowcolor{Gray} \multicolumn{1}{c|}{Methods ($\downarrow$)} & Lab & Unlab & \multicolumn{1}{c|}{\cellcolor[HTML]{ECF4FF}All} & Lab & Unlab & \multicolumn{1}{c|}{\cellcolor[HTML]{ECF4FF}All} & Lab & Unlab & \multicolumn{1}{c|}{\cellcolor[HTML]{ECF4FF}All} & Lab & Unlab & \multicolumn{1}{c|}{\cellcolor[HTML]{ECF4FF}All} & Lab & Unlab & \cellcolor[HTML]{ECF4FF}All \\ \midrule
\rowcolor{Gray} \multicolumn{16}{c}{Task Aware Evaluation} \\ \midrule
\multicolumn{1}{l|}{RS \cite{han2019automatically}} & 20.00 & 84.48 & \multicolumn{1}{c|}{\cellcolor[HTML]{ECF4FF}52.24} & 44.1 & 55.7 & \multicolumn{1}{c|}{\cellcolor[HTML]{ECF4FF}49.9} & 18.14 & 32.56 & \multicolumn{1}{c|}{\cellcolor[HTML]{ECF4FF}25.35} & 13.05 & 11.5 & \multicolumn{1}{c|}{\cellcolor[HTML]{ECF4FF}12.28} &  3.34 & 24.54  & 13.94 \cellcolor[HTML]{ECF4FF} \\
\multicolumn{1}{l|}{NCL  \cite{zhong2021neighborhood}} & 20.00 & 59.96 & \multicolumn{1}{c|}{\cellcolor[HTML]{ECF4FF}39.98} & 13.59 & 57.9 & \multicolumn{1}{c|}{\cellcolor[HTML]{ECF4FF}35.75} & 10.14 & 12.18 & \multicolumn{1}{c|}{\cellcolor[HTML]{ECF4FF}11.16} & 12.65 & 4.73 & \multicolumn{1}{c|}{\cellcolor[HTML]{ECF4FF}8.69} & 1.52 & 11.45 & 6.49 \cellcolor[HTML]{ECF4FF} \\
\multicolumn{1}{l|}{UNO \cite{Fini2021unified}} & 33.16 & \textbf{93.22} & \multicolumn{1}{c|}{\cellcolor[HTML]{ECF4FF}63.19} & 2.01 & 72.78 & \multicolumn{1}{c|}{\cellcolor[HTML]{ECF4FF}37.39} & 1.76 & 53.85 & \multicolumn{1}{c|}{\cellcolor[HTML]{ECF4FF}27.81} & 7.95 & 48.7 & \multicolumn{1}{c|}{\cellcolor[HTML]{ECF4FF}28.33} & 0.75 & 63.4 & \cellcolor[HTML]{ECF4FF}32.08 \\
\multicolumn{1}{l|}{Ours} & \textbf{92.72} & 90.32 & \multicolumn{1}{c|}{\cellcolor[HTML]{ECF4FF}\textbf{91.52}} & \textbf{65.03} & \textbf{77.03} & \multicolumn{1}{c|}{\cellcolor[HTML]{ECF4FF}\textbf{71.03}} & \textbf{73.18} & \textbf{55.66} & \multicolumn{1}{c|}{\cellcolor[HTML]{ECF4FF}\textbf{64.42}} & \textbf{84.8} & \textbf{49.67} & \multicolumn{1}{c|}{\cellcolor[HTML]{ECF4FF}\textbf{67.24}} & \textbf{27.46} & \textbf{79.07} & \cellcolor[HTML]{ECF4FF}\textbf{53.27} \\ \midrule
\rowcolor{Gray} \multicolumn{16}{c}{Generalized Evaluation} \\ \midrule
\multicolumn{1}{l|}{UNO \cite{Fini2021unified}} & 0 & 71.36 & \multicolumn{1}{c|}{\cellcolor[HTML]{ECF4FF}35.68} & 0 & 58.15 & \multicolumn{1}{c|}{\cellcolor[HTML]{ECF4FF}29.08} & 0 & 34.22 & \multicolumn{1}{c|}{\cellcolor[HTML]{ECF4FF}17.11} & 0 & 41.61 & \multicolumn{1}{c|}{\cellcolor[HTML]{ECF4FF}20.81} & 0 & 68.34  & \cellcolor[HTML]{ECF4FF} 34.17 \\
\multicolumn{1}{l|}{Ours} & \textbf{79.68} & \textbf{73.66} & \multicolumn{1}{c|}{\cellcolor[HTML]{ECF4FF}\textbf{76.67}} & \textbf{53.23} & \textbf{60.6} & \multicolumn{1}{c|}{\cellcolor[HTML]{ECF4FF}\textbf{56.92}} & \textbf{62.76} & \textbf{36.42} & \multicolumn{1}{c|}{\cellcolor[HTML]{ECF4FF}\textbf{49.59}} & \textbf{57.85} & \textbf{42.18} & \multicolumn{1}{c|}{\cellcolor[HTML]{ECF4FF}\textbf{50.02}} & \textbf{21.32} &  \textbf{70.99} & \cellcolor[HTML]{ECF4FF} \textbf{46.16} \\ \bottomrule
\end{tabular}%
}
% \vspace{-27pt}
\end{table}

% \vspace{-2pt}
\subsection{ Results}\label{sec:icd_results}
We summarise our main results in \Cref{tab:main-results}. In the first row, we refer to the different data splits via the following concise notation: dataset$-$total\_class\_count$-$lab
eled\_classes$-$unlabeled\_classes. 
`Lab' and `Unlab' columns refer to the performance of the model on the labeled and the unlabeled data respectively, after learning to discover novel categories. `All' column gives the average performance  which gives a holistic measure of capacity across all classes.
The first section of the table showcases the results in a task-aware setting. RS, NCL and UNO tend to forget how to detect instances from the labeled classes while trying to discover novel categories from unlabeled data. The unified head approach in UNO substantially improves the performance of class discovery. Our proposed pseudo-latent based replay mechanism, combined with MI based regularization helps to achieve improved class discovery capability while retaining the performance on the labeled classes. The forgetting is even more intense in the task-agnostic evaluation setting due to the inherent confusion caused due to absence of task identifying information. \LIIshort helps to address this to an extent, which complements the improved performance of all the classes, when compared to the baseline. 

On CIFAR-100 dataset, we experiment with changing the ratio of the labeled and unlabeled classes. We see a steady decrease in the class discovery performance and an increase in performance in recognizing instances from labeled classes when there are lesser number of classes in the labeled pool. This implies that better pertaining on more variety of labeled classes will improve NCD.

\section{Analysis}
We complement our main results with additional analysis on our methodological components, ablation results and measuring the sensitivity to hyper-parameters.

\subsection{Learning the \LII}\label{sec:trainingLLI}

\begin{figure}[h]
\vspace{-23pt}
\includegraphics[width=\columnwidth]{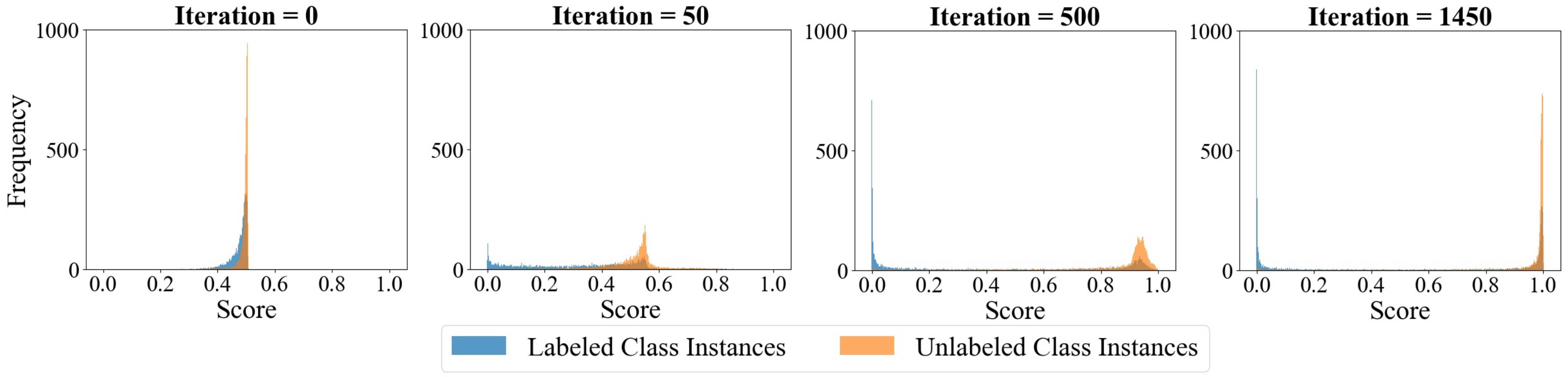}
\vspace{-20pt}
\caption{\LIIshort learns to discriminate between latent representations from unlabeled data and the pseudo latents. These plots show the classification performance on test set of labeled and unlabeled classes, showing how \LIIshort generalizes as the learning progresses.
}
\label{fig:learning_LII}
\vspace{-10pt}
\end{figure}

\begin{wraptable}[7]{r}{0.6\textwidth}
\vspace{-25pt}
\centering
\caption{Sensitivity analysis on the threshold $\tau$.}
\vspace{-10pt}
\label{tab:sensitivityTau}
\resizebox{0.6\textwidth}{!}{%
\begin{tabular}{>{\kern-\tabcolsep}c|ccc|ccc|ccc|ccc<{\kern-\tabcolsep}}
\toprule
\rowcolor{Gray} Setting & \multicolumn{3}{c|}{CIFAR-10-5-5} & \multicolumn{3}{c|}{CIFAR-100-80-20} & \multicolumn{3}{c|}{CIFAR-100-50-50} & \multicolumn{3}{c}{CIFAR-100-20-80} \\ \midrule
\rowcolor{Gray} $\tau$ & Lab & Unlab & \cellcolor[HTML]{ECF4FF}All & Lab & Unlab & \cellcolor[HTML]{ECF4FF}All & Lab & Unlab & \cellcolor[HTML]{ECF4FF}All & Lab & Unlab & \cellcolor[HTML]{ECF4FF}All \\ \midrule
0.8 & 64.94 & 80.66 & \cellcolor[HTML]{ECF4FF}72.8 & 47.85 & 70.51 & \cellcolor[HTML]{ECF4FF}59.18 & 47.85 & 51.95 & \cellcolor[HTML]{ECF4FF}49.9 & 42.87 & 48.73 & \cellcolor[HTML]{ECF4FF}45.8 \\
0.85 & 66.21 & 79.58 & \cellcolor[HTML]{ECF4FF}72.9 & 48.73 & 69.05 & \cellcolor[HTML]{ECF4FF}58.89 & 48.63 & 51.66 & \cellcolor[HTML]{ECF4FF}50.15 & 44.33 & 48.43 & \cellcolor[HTML]{ECF4FF}46.38 \\
0.9 & 69.53 & 77.83 & \cellcolor[HTML]{ECF4FF}73.68 & 49.7 & 69.14 & \cellcolor[HTML]{ECF4FF}59.42 & 50.39 & 50.87 & \cellcolor[HTML]{ECF4FF}50.63 & 45.41 & 47.94 & \cellcolor[HTML]{ECF4FF}46.68 \\
0.95 & 72.55 & 74.91 & \cellcolor[HTML]{ECF4FF}73.73 & 52.24 & 66.99 & \cellcolor[HTML]{ECF4FF}59.62 & 53.32 & 49.02 & \cellcolor[HTML]{ECF4FF}51.17 & 49.31 & 46.77 & \cellcolor[HTML]{ECF4FF}48.04 \\
\rowcolor{highlight}0.99 & 79.68 & 73.66 & \cellcolor[HTML]{ECF4FF}76.67 & 53.23 & 60.6 & \cellcolor[HTML]{ECF4FF}56.92 & 62.76 & 36.42 & \cellcolor[HTML]{ECF4FF}49.59 & 57.85 & 42.18 & \cellcolor[HTML]{ECF4FF}50.02 \\
0.999 & 85.05 & 51.66 & \cellcolor[HTML]{ECF4FF}68.36 & 61.42 & 45.5 & \cellcolor[HTML]{ECF4FF}53.46 & 69.23 & 27.14 & \cellcolor[HTML]{ECF4FF}48.19 & 65.62 & 34.17 & \cellcolor[HTML]{ECF4FF}49.9 \\ \bottomrule
\end{tabular}%
\vspace{-25pt}
}

\end{wraptable}

In \Cref{fig:learning_LII} we visualise how the \LII matures as the learning progresses in the CIFAR-100-50-50 setting. Before the learning starts, both the labeled and unlabeled latents are classified equally-likely. As the learning progresses, the \LIIshort is able to disambiguate the majority of the labeled and unlabeled samples. Still, there are some false-positives which is the reason for the performance difference between task-aware and generalized evaluation in \Cref{tab:main-results}. We run a sensitivity analysis on $\tau$ (the threshold used to decide on the prediction from \LIIshort) in \Cref{tab:sensitivityTau}. As $\tau$ increases, the performance on the labeled data increases and the unlabeled performance decreases. We use $\tau = 0.99$ throughout our experiments.

\subsection{Ablation Experiments}
We systematically remove the two ingredients in our methodology and report the results in \Cref{tab:ablation}. We evaluate with CIFAR-10 and CIFAR-100-20-80 settings. The 20-80 setting is the most practical among the various setting on CIFAR-100 as it has lower number of labeled classes, and more unlabeled classes. 
\begin{wraptable}{r}{0.5\textwidth}
% \vspace{-4pt}
\centering
\caption{Results on ablating Pseudo Latent Replay (PLR) and Mutual Information Regularizer (MIR) from our methodology.}
\vspace{-5pt}
\label{tab:ablation}
\resizebox{0.5\textwidth}{!}{%
\begin{tabular}{>{\kern-\tabcolsep}lcc|ccc|ccc<{\kern-\tabcolsep}}
\toprule
\rowcolor{Gray} \multicolumn{3}{c|}{Settings} & \multicolumn{3}{c|}{CIFAR-10-5-5} & \multicolumn{3}{c}{CIFAR-100-20-80} \\ \midrule
\rowcolor{Gray} \multicolumn{1}{l|}{} & \multicolumn{1}{l|}{PLR} & MIR & Lab & Unlab & \cellcolor[HTML]{ECF4FF}All & Lab & Unlab & \cellcolor[HTML]{ECF4FF}All \\ \midrule
\multicolumn{1}{l|}{Upper-bound} & \multicolumn{1}{c|}{-} & - & 98.24 & 94.15 & \cellcolor[HTML]{ECF4FF}96.2 & 88.85 & 53.71 & \cellcolor[HTML]{ECF4FF}71.28 \\ \midrule
\multicolumn{1}{l|}{\multirow{3}{*}{Ours}} & \multicolumn{1}{c|}{\checkmark} & $\times$ & 92.12 & 85.57 & \cellcolor[HTML]{ECF4FF}88.85 & 84.95 & 43.34 & \cellcolor[HTML]{ECF4FF}64.15 \\
\multicolumn{1}{l|}{} & \multicolumn{1}{c|}{$\times$} & \checkmark & 34.12 & 91.08 & \cellcolor[HTML]{ECF4FF}62.6 & 6.75 & 50.97 & \cellcolor[HTML]{ECF4FF}28.86 \\
\multicolumn{1}{l|}{} & \multicolumn{1}{c|}{\checkmark} & \checkmark & 92.72 & 90.32 & \cellcolor[HTML]{ECF4FF}91.52 & 84.8 & 49.67 & \cellcolor[HTML]{ECF4FF}67.24 \\ \bottomrule
\vspace{-40pt}
\end{tabular}%
}
\end{wraptable}
We note that removing the pseudo-latent replay (PLR, \Cref{sec:alleviating_forgetting}), causes significant catastrophic forgetting. Selectively turning off the mutual-information based regularizer (MIR, \Cref{sec:enhancing_class_discovery}) reduces the performance of class discovery. These results brings out the efficacy of our constituent methodological components. We note that our proposed method significantly closes the gap with the upper-bound which has access to the labeled and unlabeled data together, during training. 

\subsection{On Mutual Information based Regularization} \label{sec:misclassification}
\begin{figure}[h]
\vspace{-10pt}
\includegraphics[width=\columnwidth]{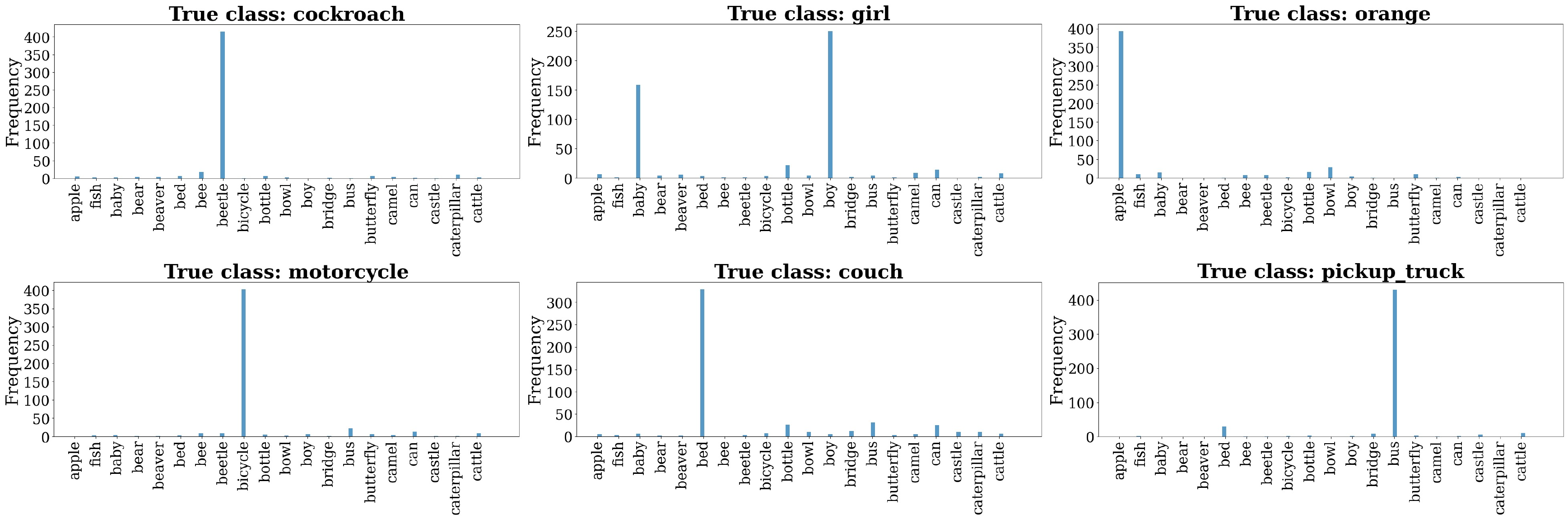}
\vspace{-10pt}
\caption{The $x$-axis in these frequency plots represents the labeled classes in CIFAR-100-20-80 setting. Each plot shows predictions for instances of an unlabeled class (referred to as `True class') from the labeled head. We see that most of the unlabeled instances gets misclassified into semantically meaningful labeled categories.
}
\label{fig:mihelps}
\vspace{-5pt}
\end{figure}

As illustrated in \Cref{fig:mihelps}, an unlabeled instance gets misclassified into a semantically similar labeled category.
This motivates us to enhance class discovery using this semantic information in the head trained on the labeled data. We couple 
\begin{wraptable}[7]{r}{0.45\textwidth}
\vspace{-17pt}
\centering
\caption{Adding Mutual Information Regularizer (MIR) to standard NCD method UNO \cite{Fini2021unified}.}
\vspace{-5pt}
\label{tab:addingToUNO}
\resizebox{0.45\textwidth}{!}{%
\begin{tabular}{>{\kern-\tabcolsep}l|ccc<{\kern-\tabcolsep}}
\toprule
\rowcolor{Gray} Setting & CIFAR-10-5 & CIFAR-100-80 & CIFAR-100-50 \\ \midrule
UNO & 94.15 & 89.31 & 59.45 \\
UNO + MIR & \textbf{94.43} & \textbf{91.26} & \textbf{61.23} \\ \bottomrule
\end{tabular}%
}
% \vspace{-15pt}
\end{wraptable}
the mutual dependency between  labeled and unlabeled heads by maximizing the mutual information between them. This helps to transfer the semantic information from the labeled to the unlabeled head, effectively guiding its class discovery capability.
Such improvement is evident from the results in \Cref{tab:main-results} 
and \Cref{tab:ablation}.
Further, we validate the efficacy of maximizing the mutual-information between the labeled and unlabeled head in standard NCD setting too by adding it to UNO \cite{Fini2021unified}. Our extra regularization is able to positively improve in this setting too, as seen in the results in \Cref{tab:addingToUNO}.

% \begin{minipage}{\textwidth}
%   \begin{minipage}[b]{0.70\textwidth}
%     \centering
%     \includegraphics[width=\columnwidth]{images/oc.pdf}
%     \captionof{figure}{A table beside a figure}
%   \end{minipage}
%   \hfill
%   \begin{minipage}[b]{0.27\textwidth}
%     \centering
% % \caption{Adding MIR}
% \label{tab:addingMIR}
% \resizebox{\textwidth}{!}{%
% \begin{tabular}{@{}l|cc@{}}
% \toprule
% \multicolumn{1}{c|}{Setting} & UNO & UNO + MIR \\ \midrule
% CIFAR-10-5-5 & 94.15 & 94.43 \\
% CIFAR-100-80-20 & 89.31 & 91.26 \\
% CIFAR-100-50-50 & 59.45 & 61.23 \\ \bottomrule
% \end{tabular}%
% }
%  \captionof{table}{A table beside a figure}
%     \end{minipage}
%   \end{minipage}

% \vspace{10pt}
\subsection{Sensitivity Analysis on Mixing Coefficient}
\begin{wraptable}{r}{0.6\textwidth}
\vspace{-24pt}
\centering
\caption{The mixing coefficient $\alpha$ is sampled from $Beta(\gamma, \rho)$. We vary $\gamma, \rho$ to measure the sensitivity of our method to the degree of mixing.}
\vspace{-5pt}
\label{tab:alphaAnalysis}
\resizebox{0.6\textwidth}{!}{%
\begin{tabular}{>{\kern-\tabcolsep}cc|ccc|ccc|ccc|ccc<{\kern-\tabcolsep}}
\toprule
\rowcolor{Gray} \multicolumn{2}{c|}{$\alpha \sim Beta(\gamma, \rho)$} & \multicolumn{3}{c|}{CIFAR-10-5-5} & \multicolumn{3}{c|}{CIFAR-100-80-20} & \multicolumn{3}{c|}{CIFAR-100-50-50} & \multicolumn{3}{c}{CIFAR-100-20-80} \\ \midrule
\rowcolor{Gray} $\gamma$ & $\rho$ & Lab & Unlab & \cellcolor[HTML]{ECF4FF}All & Lab & Unlab & \cellcolor[HTML]{ECF4FF}All & Lab & Unlab & \cellcolor[HTML]{ECF4FF}All & Lab & Unlab & \cellcolor[HTML]{ECF4FF}All \\ \midrule
1 & 1 & 92.92 & 90.93 & \cellcolor[HTML]{ECF4FF}91.93 & 64.6 & 76.17 & \cellcolor[HTML]{ECF4FF}70.39 & 73.74 & 55.22 & \cellcolor[HTML]{ECF4FF}64.48 & 84.75 & 47.07 & \cellcolor[HTML]{ECF4FF}65.91 \\
100 & 1 & 92.52 & 91.31 & \cellcolor[HTML]{ECF4FF}91.92 & 64.8 & 12.76 & \cellcolor[HTML]{ECF4FF}38.78 & 74.06 & 42.95 & \cellcolor[HTML]{ECF4FF}58.51 & 84.95 & 40.33 & \cellcolor[HTML]{ECF4FF}62.64 \\
1 & 100 & 92.72 & 90.32 & \cellcolor[HTML]{ECF4FF}91.52 & 65.03 & 77.03 & \cellcolor[HTML]{ECF4FF}71.03 & 73.18 & 55.66 & \cellcolor[HTML]{ECF4FF}64.42 & 84.8 & 49.67 & \cellcolor[HTML]{ECF4FF}67.24 \\ \bottomrule
\end{tabular}%
}
\vspace{-20pt}
\end{wraptable}
The pseudo-latent representations $\bm{z}_{p}$ that we generate using \Cref{algo:replay} uses a mixing coefficient $\alpha$ to do a linear combination of the inverted latent representation $\bm{z}_{L}$ and the corresponding class mean $\bm{z}^{c}_{\mu}$: $\bm{z}_{p} = \alpha \bm{z}_{L} + (1 - \alpha) \bm{z}^{c}_{\mu}$. $\alpha$ is indeed sampled from $Beta(\gamma, \rho)$. As we increase $\gamma$, $\bm{z}_{p}$ will be closer to the inversed latents, while increasing $\rho$ the class means will have more importance. When $\gamma = \rho = 1$, we get uniform samples between $0$ and $1$. We experiment with these three configurations in \Cref{tab:alphaAnalysis}. For the simple CIFAR-10 dataset, we see negligible effect on varying $\alpha$.
On the harder CIFAR-100 dataset, we see that sampling $\alpha$ uniformly or closer to the class means helps to retain past knowledge without affecting class discovery. This result emphasises the fact that class means imparts semantic information and its interpolation with inversed latents provides diverse pseudo latents.

% \subsection{t-SNE Visualisations}

\section{Conclusion}
% We introduce the pragmatic \gcdwf setting.
% This makes use of a pre-trained image classifier to discover novel categories from unlabeled data, without degrading its performance on the base classes. Our proposed methodology makes use of pseudo-latents as a surrogate to labeled instances to defy forgetting, and a mutual-information based regularizer to enhance class discovery. We operate in a generalized setting, aided by a binary latent discriminator which segregates labeled instance from the unlabeled ones during inference. 
% % \gcdwfshort models is effective in application contexts like a robot navigating an environment or  
% % Besides being effective in applications like recognition models in robots, \gcdwfshort can enhance the practicality of other problem settings like Open World Detection \cite{bendale2015towards,joseph2021towards} and Active Learning \cite{cohn1996active}. The automatically discovered classes
% % (Incomplete, will add.)
% % A \gcdwfshort model which discovers novel classes without forgetting can 
% % \gcdwfshort is a niche problem setting which presents a galore of opportunities to 
% \gcdwfshort presents a versatile problem setting with wide practical applicability. We hope our work will kindle further interest into this under-explored area of machine learning. 

We introduce \gcdwf, a pragmatic extension to NCD setting. We develop an effective approach for \gcdwfshort, which makes use of pseudo-latents as a surrogate to labeled instances to defy forgetting, and a mutual-information based regularizer to enhance class discovery. We operate in a generalized setting, where a test instance can come from any of the classes of interest. We propose to use \LII to segregate labeled instance from the unlabeled ones during inference. We report results on five different data-splits across three datasets to test the mettle of our approach. 
% We achieve superior performance when benchmarked against three prominent NCD baselines. 
We hope our work can shed light on this challenging problem and inspire more efforts towards this realistic setting.

% During inference, a test instance may come from either old and new classes in \gcdwfshort, we introduce a \LII to identify whether an unlabeled image is from an old class or a new one. We report results on public datasets, showing superior performance over the prominent NCD baselines. We hope our work can shed light on this challenging problem and inspire more efforts towards this realistic setting.

\vspace{10pt}
\noindent \textbf{Acknowledgements:} KJJ was a Student Researcher at Google.
% KJJ thanks Google for  the Student Researcher position.
We are grateful to the Department of Science and Technology, India, as well as Intel India for the financial support of this project through the IMPRINT program (IMP/2019/000250). 
This work is supported in part by Hong Kong Research Grant Council - Early Career Scheme (Grant No. 27208022) and HKU Startup Fund.
We also thank our anonymous reviewers for their valuable feedback.
% in improving the presentation of this paper.
% We also thank the anonymous reviewers for their valuable feedback in improving the presentation of this paper.
\clearpage
\bibliographystyle{splncs04}
\bibliography{egbib}
\end{document}